%% file: norm_logic/main.tex
\pdfoutput=1

\documentclass[11pt]{article}

\usepackage[final]{acl}

\usepackage{times}
\usepackage{latexsym}

\usepackage[T1]{fontenc}
\usepackage[utf8]{inputenc}

\usepackage{microtype}

\usepackage{inconsolata}

\usepackage{graphicx}

\usepackage{booktabs}
\usepackage{amssymb}
\usepackage{multirow}
\usepackage{tcolorbox}
\usepackage{amsmath} 
\usepackage{arydshln}
\usepackage[table]{xcolor}
\usepackage{siunitx}
\usepackage{enumitem}

\title{Beyond Surface Forms: Symbolic Edits as a Test for Logical Reasoning with LLMs}

\author{
 \textbf{Ramya Keerthy Thatikonda\textsuperscript{$\forall$}} \ \ \
 \textbf{Wray Buntine\textsuperscript{$\forall$}} \ \ \
 \textbf{Ehsan Shareghi\textsuperscript{$\exists$}}
\\
\\
 \textsuperscript{$\forall$}Department of Data Science \& AI, Monash University
 \\
 \textsuperscript{$\exists$}Department of Computer Science, University College London
\\
\\
\texttt{\{ramya.thatikonda1,wray.buntine\}@monash.edu} \\
 \texttt{ehsan.shareghi@ucl.ac.uk}
}

\begin{document}
\maketitle
\begin{abstract}
Logical reasoning with large language models (LLMs) is a critical capability, as it reflects a system’s ability to correctly deduce hypotheses from a given context using faithful deductive processes. However, LLM reasoning has often been shown to be sensitive to small surface-level variations in problem formulation, raising questions about whether models truly follow the underlying logical structure. Studying this behavior is challenging because the symbolic components of logical problems, such as operators and predicates, are difficult to systematically manipulate in natural language. We introduce a tool-driven framework for generating controlled, label-preserving edits to logical reasoning problems. Our method operates on symbolic representations of first-order logic and constraint satisfaction problem tasks, enabling targeted modifications to logical operators and other structural components before translating them back into natural language. Using this framework, we evaluate various LLMs under cumulative and individual operator edits and analyze their behavior in response to these changes. Our quantitative and qualitative analyses show that LLM reasoning behavior under controlled operator edits is inconsistent, regardless of model size or family: models sometimes adapt correctly to structural changes but often fail to track their logical consequences. The results from this automated stress test enable an evaluation of language models across different dimensions and help measure the reliability of their reasoning.
\footnote{The code and dataset are available at \href{https://github.com/RamyaKeerthy/BeyondSurfaceForms.git}{RamyaKeerthy/BeyondSurfaceForms}.}
\end{abstract}

\begin{figure}
    \centering
    \includegraphics[width=\columnwidth, clip]{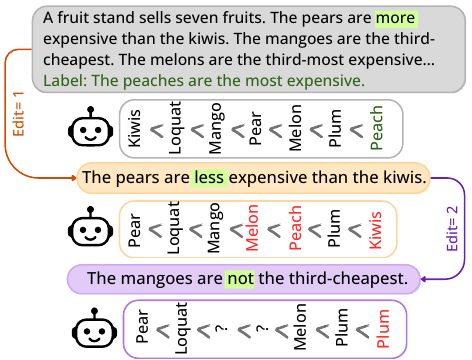}
    \caption{Example from the \texttt{Deduction} dataset, with outputs generated by the \texttt{Gemma3-27B} model, showing how label-preserving edits disrupt the model’s reasoning. For instance, at edit level 1, the operator changes from \textit{Pears > Kiwi} to \textit{Pears < Kiwi}, leading the model to incorrectly interpret the final order.
    }
    \label{fig:error-exampleNew}
    \end{figure}

\section{Introduction}
\input{norm_logic/Introduction}
\label{sec:intro}

\section{Related Work}
\input{norm_logic/RelatedWork}
\label{sec:rw}

\section{Methodology}
\input{norm_logic/NormLogic}
\label{sec:methodology}

\section{Experiments}
\input{norm_logic/Experiments}
\label{sec:experiments}

\section{Results and Discussion}
\input{norm_logic/Discussion}
\label{sec:discussion}

\section{Conclusion}
\input{norm_logic/Conclusion}
\label{sec:conclusion}

\section*{Limitations}

Our work is subject to the following limitations.

\begin{itemize}

    \item \textbf{Tool dependency and translation coverage.} The methodology is contingent on the availability of translations supported by the tool used. When translations are unavailable for a given problem, the analysis cannot be performed as-is, since automated translation forms the foundation of our approach. Our study uses a multiple pass at translation for logical deduction and ar-lsat datasets. We rely on availability of parsable translations for datasets such as FOLIO, which limits the number of translations. Furthermore, even where translations are available, semantic errors may remain latent; these may not surface during tool-based inference if they do not affect the underlying reasoning chain.
    
    \item \textbf{Back-translation fidelity.} The back-translation from first-order logic to natural language is designed to be as semantically transparent as possible, with explicit constraints provided to the language model. Although human evaluation revealed minimal errors, the non-deterministic nature of language models (we use Gemini 2.5 Flash) may introduce translation errors, particularly for complex logical expressions. Errors introduced during the initial natural-language-to-symbolic translation may also propagate to the subsequent back-translation. As our work is evaluation-based and measures model reasoning ability after back-translation, we expect such errors to have a minimal effect on the overall results, particularly given that at least one model correctly solves these back-translated questions.

    \item \textbf{Reproducibility with proprietary models.} Our study makes use of proprietary language models, which are non-deterministic and subject to undisclosed version updates, limiting exact replication of results. To mitigate this, we report results for open-source models with explicit version identifiers, which can be used to reproduce our findings.
\end{itemize}

\section*{Acknowledgement}
This research was supported by an Australian Government Research Training Program (RTP) Scholarship, doi.org/10.82133/C42F-K220

\bibliography{norm_logic/custom}

\appendix
\input{norm_logic/Appendix}
\label{sec:appendix}

\end{document}

%% file: norm_logic/Introduction.tex
Deductive logical reasoning is an important testbed for studying the ability of large language models (LLMs) to perform clear, step-by-step inference while following explicitly specified rules. Recent work has reported performance saturation on many existing logical reasoning benchmarks \cite{proverqa, deng2024enhancing}. These results have motivated growing scrutiny of benchmark reliability, particularly in mathematical reasoning, where studies ask whether reported gains reflect robust reasoning abilities or instead arise from artifacts such as training-distribution effects, benchmark contamination, or leakage \cite{mirzadeh2024gsm, stolfo2023causal}.

A common paradigm for conducting such studies is to create new versions of problems in existing benchmarks by replacing numerical values and entities, thereby altering the surface form of the question. While effective for mathematics, where such perturbations often involve simple numerical substitutions \cite{mirzadeh2024gsm}, logical reasoning depends on structured symbolic components, including operators, predicates, constants, and constraints, that collectively determine the validity of a deduction. Small changes to these components can fundamentally alter both the reasoning process and the outcome. Understanding whether LLMs respond appropriately to such structural changes is therefore critical for evaluating their capabilities. However, systematically studying this behavior remains difficult. Existing evaluations rely on manually created problem variations, which are costly to produce and difficult to scale \cite{stolfo2023causal}. To the best of our knowledge, there is currently no systematic framework for generating controlled modifications to logical reasoning problems that isolate structural changes while tracking their effects on the final outcome.

To address this challenge, we introduce a structural analysis framework for logical reasoning that uses tool-driven template generation to produce controlled, label-preserving variations of logical deduction problems. The framework decomposes textual reasoning problems into their underlying logical components using symbolic representations and enables targeted interventions on specific elements of the logical structure using a symbolic solver as a tool. We specifically focus on operator-level edits, allowing us to examine how model performance changes when logical operators are modified while tracking changes in the final answer. Figure~\ref{fig:error-exampleNew} shows an example of how a change in an operator affects the model's performance in determining the final answer.

We conduct a comprehensive operator-edit analysis along two dimensions: cumulative edits that preserve the final answer and single edits that can change the ground truth of the problem. We initially evaluate these interventions using final-answer accuracy to capture changes in predictions under label-preserving edits. We measure the performance of several LLMs on deductive logical reasoning and constraint satisfaction tasks, two categories that require reasoning from logical structure rather than external world knowledge. We follow this by applying single edits that can modify the ground-truth answer, allowing us to assess whether the models reliably adapt their final predictions to these changes. Our experiments stress-test different natural-language reasoning strategies, including standard chain-of-thought prompting \cite{wei2022chain}, inference-time scaling \cite{snell2024scaling, wu2024inference}, and symbolic chain-of-thought reasoning \cite{xu2024faithful}.

Overall, our study provides a systematic analysis of how LLM reasoning behavior changes under controlled structural interventions. By generating variations of logical reasoning problems, we enable evaluation beyond traditional final-answer accuracy and examine whether models consistently track the logical consequences of these structural edits. Our results across recent LLMs with different reasoning strategies show that model behavior under operator interventions is inconsistent: reasoning changes rapidly under logical modifications and varies unpredictably. Label-changing edits further help us understand the extent to which models fall back on known answers, indicating potential reliance on previously learned answer patterns. In addition, an LLM-based qualitative analysis examines reasoning paths and provides insight into the conditions that lead a drop in reasoning stability. These findings suggest that current language models do not reliably ground their reasoning in the symbolic structure of the problem, underscoring the need for benchmarks that explicitly probe structural reasoning in LLMs.

%% file: norm_logic/RelatedWork.tex
\paragraph{Analysis of LLM Behavior}
Understanding how structural variations in input affect language model behavior has long been a challenging problem \cite{liao2023ai}. Prior work has explored memorization in LLMs through perturbation-based analyses \cite{xie2025memorization}. While such approaches provide insights into model sensitivity, they do not explicitly control for structural factors across perturbations. In contrast, our work focuses on manipulations at the operator level, enabling a more fine-grained analysis of structural effects.

In logical deduction and stress testing, prior studies have examined how models behave under changes to premise order and paraphrasing \cite{chen2024premise,bao2025less}. Other works in mathematical reasoning have used irrelevant context to demonstrate model performance degradation \cite{shi2023large,yang2025llm}. We follow a similar direction in highlighting current gaps in LLM evaluation, but focus on automated template generation to create controlled logic-form variations grounded in symbolic structure.

\paragraph{Surface Form Interventions}
Recent work in mathematical reasoning has explored controlled interventions to study model behavior. \citet{stolfo2023causal} demonstrate the feasibility of causal interventions in mathematical reasoning tasks, although their approach is primarily limited to value-level edits. Similarly, \citet{mirzadeh2024gsm} examine how variations in numerical values within datasets such as GSM8K influence model performance. Other data generation strategies include counterfactual task generation \cite{lewis2024using}, where counterfactual variants of existing analogy problems are constructed to test the reasoning abilities, and metamorphic testing \cite{cho2025metamorphic}, where predefined metamorphic relations are used to generate related test inputs. These approaches further demonstrate how variations of existing benchmarks can be used to examine LLM performance. While these works provide valuable insights into how LLMs respond to changes in input values or task distributions, such approaches can require costly human annotation or dataset construction. In contrast, we propose a symbolic approach to systematically generate such variations with minimal manual effort.

\begin{figure*}
    \centering
    \includegraphics[width=\textwidth]{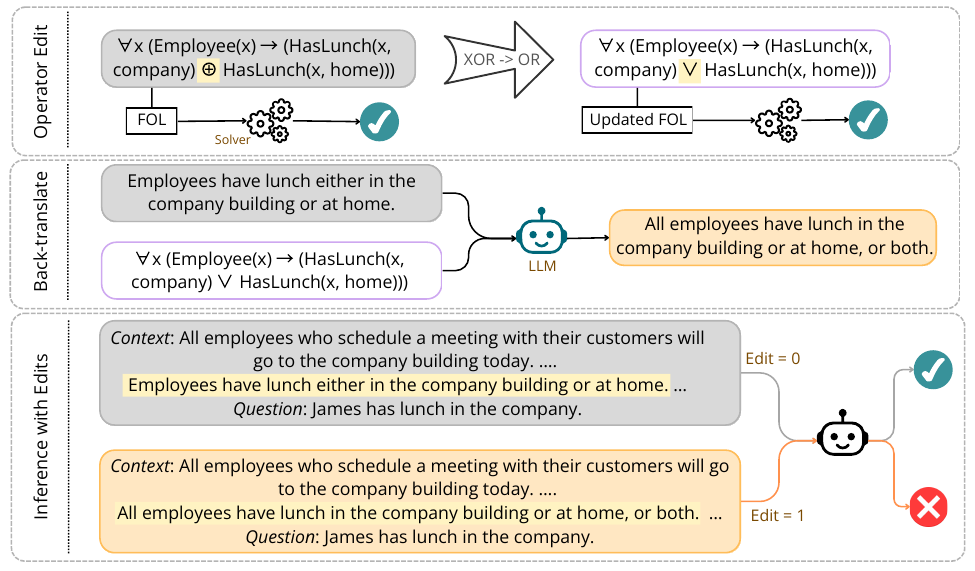}
    \caption{Overview of logic-form edits data extraction. The first step comprises operator edits performed by the tool wrapper, which can be label preserving or altering verified using a solver. In the second step, back-translation by the LLM is conditioned on the valid edited symbolic representation along with the original text. At inference, a second LLM is evaluated to produce outputs that demonstrate the effect of structural changes on LLM performance.}
    \label{fig:method}
    \end{figure*}

\paragraph{Reasoning with Deductive Logic}
Recent surveys highlight the importance of logical reasoning for LLMs \cite{cheng2025empowering,liu2025logical}. Prior work has studied symbolic generation \cite{pan2023logic}, natural-language reasoning \cite{wei2022chain,thatikonda2025logical}, and hybrid approaches that avoid external tools \cite{xu2024faithful}. We focus on natural-language reasoning to evaluate LLM reasoning behavior rather than translation ability.

\vspace{-2mm}

%% file: norm_logic/NormLogic.tex
\subsection{Controlled Intervention}
Given the symbolic translations, we intervene on the representations to introduce controlled variations into the questions, using a theorem prover as a wrapper. The types of interventions depend on the structure of the symbolic form. First-order logic representations consist of predicates, operators, and variables, whereas constraint satisfaction problems comprise variables, operators, and values. We focus on modifying operators in both formulations by using symbolic tools. Because such interventions are difficult to apply reliably in natural language, we operate directly on the symbolic representations, which enable precise identification and tracking of edit locations. We perform sentence interventions in two stages: cumulative edits and individual edits.

\subsubsection{Cumulative Edits}
In the first stage of intervention, we modify the input questions without altering their underlying final deductive outcomes. A question may undergo zero or more edits, depending on the availability of valid operator modifications. Each sentence in a question is evaluated to determine whether a valid modification exists. An operator edit is considered valid if the final deductive outcome remains unchanged. For example, if the constraint “the blue book is to the right of the yellow book” is changed to “the blue book is to the left of the yellow book,” the final solution must remain invariant. Regardless of whether an edit is applied, the process proceeds to the next sentence, and previously modified statements are not revisited. Subsequent edits operate on the already modified sentences, progressively increasing their divergence from the original structure. As the number of edits increases, the structural similarity to the original question decreases. These interventions enable a systematic assessment of model robustness by testing whether reasoning outcomes remain invariant under label-preserving transformations. 

Due to the cumulative nature of these edits, we use the number of edits, $k$, as the variable for measuring performance. Since not all questions admit the same number of valid edits, we use normalized accuracy, computed over $N$ records at edit level $k$. If a record does not have a $k$-th edit, its baseline prediction is used instead.
\[
\mathrm{Acc_{norm}}(k)
=
\frac{1}{N}
\sum_{i=1}^{N}
\mathbf{1}\!\left[
\hat{y}_{i,\tilde{k}} = y_i
\right],
\]
\vspace{-1mm}
\[
\qquad
\tilde{k}=
\begin{cases} k, & \text{if an edit is available at } k,\\ 0, & \text{otherwise}. \end{cases}
\]
where $y_i$ is the ground-truth deductive outcome for input $i$, and $y_{i,k}$ is the model’s prediction after $k$ edits.
This form of evaluation helps measure accuracy changes relative to the baseline (cumulative edits), rather than relative to the preceding edit (consecutive edits).

\subsubsection{Individual Edits}
In the second stage, we apply single alterations that either preserve or change the final outcome. Here, symbolic formulations are handled one at a time without accumulating changes. For interventions that change the final outcome, new ground-truth labels are stored for each question. We consider two conditions: label-altering and label-preserving. In both cases, the tool wrapper stores valid symbolic representations that either change the final answer or leave it unchanged. Comparing the two conditions helps us assess the extent of potential benchmark familiarity and evaluate how well models adapt to familiar questions with different final answers. Since each data point $i$ can have multiple single-edit variations, performance is measured by aggregating results at the question level:
\[
\mathrm{Acc}_{c}
=
\frac{1}{|\mathcal{N}_{c}|}
\sum_{i \in \mathcal{N}_{c}}
\left(
\frac{1}{|\mathcal{R}_{i}|}
\sum_{r \in \mathcal{R}_{i}}
\mathbf{1}\!\left[
\hat{y}_{r}^{(c)} = y_{r}^{(c)}
\right]
\right),
\]
\vspace{-1mm}
\[
y_{r}^{(c)}
=
\begin{cases}
y_i, & c=\mathrm{preserved\_label},\\
y_{r}^{\mathrm{new}}, & c=\mathrm{altered\_label}.
\end{cases}
\]
where $R_i$ represents the set of records associated with $i$, and $N_c$ denotes the set of questions that have valid edits under condition $c$. The label $y_{r}^{(c)}$ depends on the condition $c$ to use original label $y_{i}$	or updated ground-truth label $y_{r}^{\mathrm{new}}$.

\subsection{Back-Translations}
Given the altered symbolic forms, the final step is to translate them back into natural-language statements. To reduce complexity and ensure accuracy, we perform back-translation at the sentence level. We use an LLM with a sufficient number of in-context examples to perform back-translation from grounded semantic structures, using the original statements as references. Each translation prompt includes both the altered symbolic form and the original symbolic–-text pair, which helps discourage unintended semantic drift. The prompts are designed so that the LLM treats the task as a correction task rather than pure back-translation: we provide the new logical statement and the original sentence, and ask the LLM to revise the sentence to match the new logical statement. For example, to back-translate $blue\_book < yellow\_book$, we provide the reference sentence `The blue book is to the right of the yellow book' to encourage the model to preserve the same sentence structure and return `The blue book is to the left of the yellow book'. After generating all symbolic to natural-language translations, we replace the original statements with their altered counterparts. Statements that cannot be matched to their original counterparts may result from errors in the initial translation and are therefore excluded, helping to eliminate potentially induced errors.

Figure~\ref{fig:method} shows the complete process for generating structural changes. We additionally provide a Python library to run this framework seamlessly. More details are available in Appendix~\ref{app:logicedit}.

%% file: norm_logic/Experiments.tex
\subsection{Symbolic Data Extraction}
We use three test datasets from \cite{pan2023logic} to evaluate the LLMs: FOLIO \cite{folio}, Logical Deduction (Deduction) \cite{logicaldeduction}, and AR-LSAT \cite{zhong2022analytical}. FOLIO is a first-order logic (FOL) benchmark solved using the Prover9 theorem prover~\cite{mccune2005release}; Logical Deduction consists of constraint satisfaction problems (CSPs) solved using a Python-based constraint solver; and AR-LSAT problems are solved using the Z3 solver~\cite{de2008z3}. For details on translation generation, see Appendix~\ref{app:translations}.

\subsection{Edits and Translation}
\paragraph{Operator edits.}
For datasets with first-order logic representations, such as FOLIO and AR-LSAT, we apply controlled edits to logical operators using the following mappings:
\[
\texttt{or} \rightarrow \texttt{and}, \quad
\texttt{xor} \leftrightarrow \texttt{or}, \quad
\texttt{not} \rightarrow \varnothing
\]
These edits introduce structured perturbations that modify the logical form while preserving the final deductive outcome. Additionally, for AR-LSAT, we include the mapping $\texttt{and} \rightarrow \texttt{or}$. We avoid this edit in FOLIO, because conjunctions are often used to define properties or states of an object. For example, a statement such as $\mathrm{Game}(x) \land \mathrm{LegendOfZelda}(x)$ indicates that \textit{The Legend of Zelda} is a game; replacing $\land$ with $\lor$ would make the logical statement semantically ill-formed or substantially alter its meaning. Similarly, we avoid other operator edits that could affect the inherent meaning of a sentence rather than introduce a controlled structural variation.

Edits are applied iteratively. For each question, we scan FOL premises sequentially, apply a single operator substitution, and then pass the modified problem to the theorem prover. To avoid cycling between complementary edits, we proceed to the next premise after each successful modification and continue until no further valid edits can be identified.

For CSP, we apply an analogous procedure to constraints rather than premises, using the following operator swaps:
\[
\texttt{lessthan} \leftrightarrow \texttt{greaterthan}, \quad
\texttt{equal} \leftrightarrow \texttt{notequal}.
\]

An edit is considered valid if the corresponding CSP or FOL solver returns a value from the solution set. The number of validated operator edits per question is recorded separately for each symbolic representation.
\vspace{-1mm}

\paragraph{Back-translations}
Altered symbolic forms are translated back into natural language using \texttt{Gemini 2.5 Flash} with task-specific instructions and five in-context examples per dataset. The input consists of (i) the original natural-language statement and (ii) the altered symbolic form. The model is instructed to minimally edit the statement to reflect the modified symbolic representation, preserving sentence structure and style. The resulting sentences replace the originals, yielding an expanded evaluation set (see Appendix~\ref{app:datastat}) with an additional \emph{edit count} parameter for analysis. To verify the validity of the back-translations generated using this method, we perform an additional human evaluation on a sample of records and find a high level of agreement with the generated translations. Further details on the prompts and human evaluation are provided in Appendix~\ref{app:bt}.
 
\subsection{Reasoning Inference}
We evaluate reasoning performance under cumulative edits using Chain-of-Thought (CoT), CoT with test-time scaling (Scaling), and symbolic CoT (SymbCoT) prompting across the \texttt{Gemma3} family (1B--27B)~\cite{team2025gemma}, \texttt{LLaMA3-8B}~\cite{grattafiori2024llama}, \texttt{Phi-4-Mini}~\cite{abouelenin2025phi}, and \texttt{Qwen3-4B}~\cite{qwen3technicalreport}. For individual edits, we use CoT to analyze model behavior. For prompting details and configurations, see Appendix~\ref{app:reasonprompt}.

%% file: norm_logic/Discussion.tex
\input{norm_logic/transition-small}

\subsection{Cumulative Edits}
As the maximum number of valid edits ($k$) varies across logical questions, records without $k$ valid edits fall back to their baseline predictions ($k=0$), ensuring that accuracy is normalized with respect to the total number of questions. Thus, the accuracy at $k$ is compared with the baseline ($k=0$) and captures a subset of edits at a given point. Table~\ref{tab:backfilled_accuracy} shows a representative snippet of these results, indicating minimal impact, with most configurations remaining stable across edit levels. These preliminary findings suggest that operator edits neither systematically harm nor improve model reasoning. Additionally, across the three reasoning methods, SymbCoT often underperforms relative to the other prompting strategies. The full set of results is presented in Appendix \ref{app:cuml}.

\begin{figure}[ht]
    \centering
    \includegraphics[width=\linewidth]{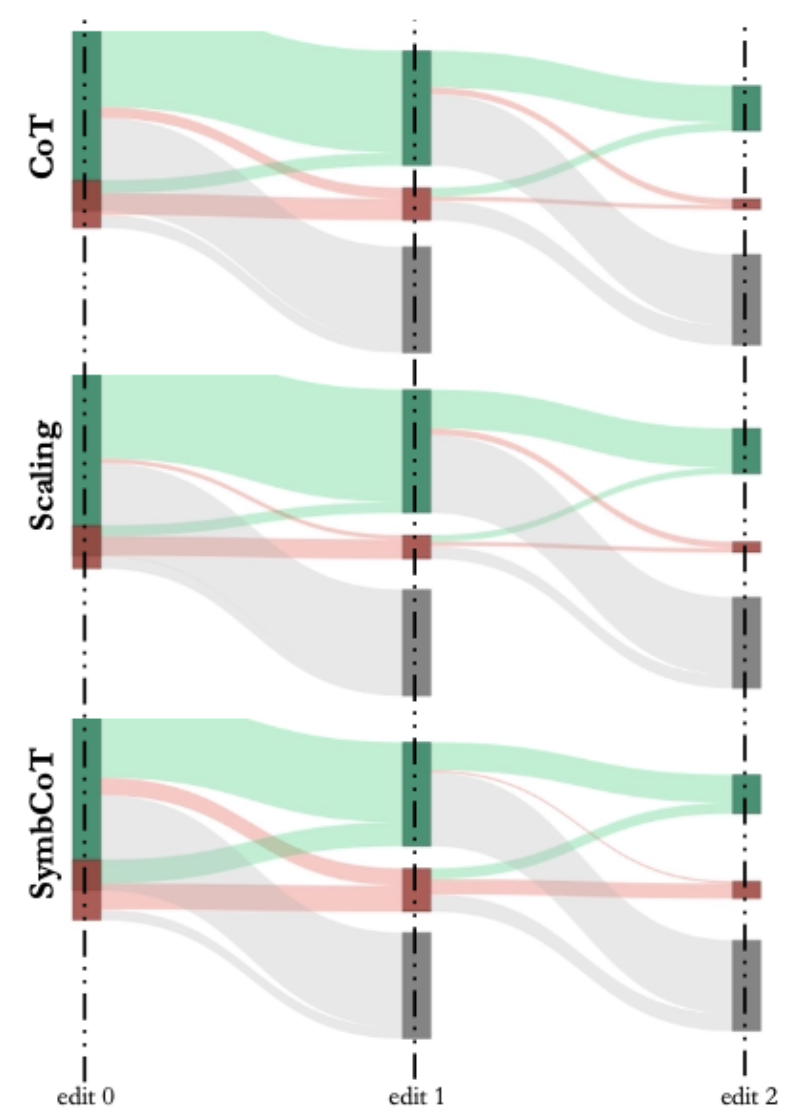}
    \vspace{-3mm}
    \caption{Transitions for FOLIO with the \texttt{Gemma3-27B} model across three forms of reasoning. Green indicates correct predictions, red indicates incorrect predictions, and grey denotes records with no further edits. Despite the insignificant change in accuracy reported in Table~\ref{tab:backfilled_accuracy}, records continue to transition between correct and incorrect predictions across edit levels.
    }
    \vspace{-3mm}
    \label{fig:tran-sankey}
\end{figure}

We observe that this minimal impact arises because transitions in record-level predictions tend to balance out in the final performance, as shown in Figure~\ref{fig:tran-sankey}. Although the large Gemma model shows stable performance in terms of accuracy, this metric does not capture the full extent of the model’s behavior. Rather than exhibiting uniform gains or drops in final accuracy, model behavior varies at the instance level as inputs undergo successive transformations. To better examine the LLMs, we analyze reasoning stability using flip rate.

\paragraph{Flip Rate}
We measure the extent to which predictions change after edits, even when the underlying semantics are preserved. Since models behave inconsistently under edit-induced changes, we analyze flips in both directions: from correct to incorrect and from incorrect to correct. We compute the overall flip rate as the number of instances whose correctness changes, divided by the total number of instances:
\[
\mathrm{FlipRate}
=
\frac{n_{\mathrm{I}\rightarrow\mathrm{C}} + n_{\mathrm{C}\rightarrow\mathrm{I}}}
{n_{\mathrm{I}\rightarrow\mathrm{C}} + n_{\mathrm{C}\rightarrow\mathrm{I}} + n_{\mathrm{I}\rightarrow\mathrm{I}} + n_{\mathrm{C}\rightarrow\mathrm{C}}},
\]
where \(n_{\mathrm{I}\rightarrow\mathrm{C}}\) denotes the number of inferences from a given LLM that change from incorrect to correct, \(n_{\mathrm{C}\rightarrow\mathrm{I}}\) denotes the number that change from correct to incorrect, and \(n_{\mathrm{I}\rightarrow\mathrm{I}}\) and \(n_{\mathrm{C}\rightarrow\mathrm{C}}\) denote inferences whose correctness remains unchanged.
\input{norm_logic/transition_table_perc}
To measure significance, we define the null hypothesis as follows: a model’s reasoning outcome remains unchanged after editing. The alternative hypothesis states that a model exhibits a significant transition in correctness with respect to the original question. We treat changes greater than (5\%) as practically meaningful and apply a Bonferroni correction when estimating (p)-values to account for multiple models for a given edit.

Table~\ref{tab:moved-pct-zero-to-k} reports the proportion of instances whose inference correctness changes after editing. We compare these results against the accuracy trends reported earlier. To better interpret the results, we analyze them by model size and dataset type.

Smaller Gemma models, particularly Gemma 1B, are highly unstable across datasets when using CoT prompting. A notable finding is that inference-time scaling substantially stabilizes smaller models on FOLIO and AR-LSAT, suggesting that this form of reasoning can benefit lower-capacity models. Among the 4B--12B models, Qwen shows lower flip rates on FOLIO and Deduction, whereas Gemma 4B shows lower flip rates on AR-LSAT. However, Qwen models also achieve the highest accuracy on AR-LSAT (Table~\ref{tab:backfilled_accuracy}), indicating that strong final-answer performance can still coexist with reasoning instability.

Gemma 27B, in contrast, shows higher stability when its overall performance is high, and lower stability when its overall accuracy is very low, as in AR-LSAT. This contrasts with the behavior of Qwen, even though the two models, despite differing substantially in size, show similar performance trends. Across the full analysis, SymbCoT exhibits mixed behavior, but overall tends to produce relatively high flip rates across datasets. This is likely due to error propagation introduced across the multiple prompting stages required to derive the final answer. We also examine whether the underlying reasoning changes across edits using reasoning length as a proxy and observe that at least one model remains stable despite changes in reasoning length (see Appendix~\ref{app:variationinproof}).

\paragraph{Qualitative study}
We conduct a qualitative study to examine the reasoning patterns produced at each edit level. We use coding criteria that assess the overall correctness of the reasoning path for logical deductive reasoning. The qualitative coding is performed across four levels, two of which evaluate the reasoning process and two of which evaluate the final answer. The criteria are designed to answer the following questions:

\begin{itemize}[leftmargin=*,nosep]
\item \textit{Premise Fidelity (PF):} Does the reasoning use only the information provided in the original question?
\item \textit{Inference Validity (IV):} Do the reasoning steps logically follow from the given premises and from prior reasoning steps?
\item \textit{Answer Completion (AC):} Does the reasoning include the necessary intermediate steps needed to support the conclusion?
\item \textit{Answer Alignment (AL):} Is the final answer supported by the reasoning?
\end{itemize}

\input{norm_logic/qual_analysis}
\input{norm_logic/singleedits}

We run the coding procedure using the \texttt{Gemini 2.5 Flash} model with a curated prompt (see Appendix~\ref{app:qualan}) designed to produce structured outputs. We apply this procedure to a subset of records across all datasets\footnote{The qualitative analysis is conducted on a small subset of records covering a large number of edits, allowing us to better study the reasoning paths. This analysis serves as a starting point to encourage further use of this form of evaluation. To the best of our knowledge, this is the first application of such an analysis in this domain, motivating future work to incorporate qualitative analysis as part of LLM evaluation.}, selecting records that contain the maximum number of edits. Each criterion receives a binary score, and the overall reasoning score is computed as the mean of these criterion-level scores. To assess the reliability of these codes, we compute the correlation between final-answer accuracy and the overall reasoning score. The two are highly correlated, with a Pearson's correlation coefficient of 0.9, providing some confidence in analyzing the individual criterion scores.

For brevity, we examine one AR-LSAT example using the \texttt{Qwen3-4B} model, where accuracy is high but stability is questionable. Table~\ref{tab:scores_by_edits} summarizes qualitative reasoning robustness under label-preserving edits. We report the overall reasoning percentage for each edit level. We observe that answer alignment, which measures whether the final answer is supported by the reasoning, decreases as question variation increases, even though accuracy does not decrease. This points to the need to examine reasoning paths more deeply rather than relying only on accuracy scores.

The model is also prompted to provide a justification whenever it assigns a score of 0 for any criterion. We analyze negatively connoted words in these justifications across all models and identify several recurring patterns that may indicate reasoning issues in LLMs. Among the frequently observed relevant tokens, we find \textit{distorts}, \textit{misinterprets}, and \textit{invents} for premise fidelity; \textit{invalid}, \textit{unsupported}, and \textit{leap} for inference validity; \textit{fails}, \textit{skips}, and \textit{incomplete} for answer completion; and \textit{flawed} and \textit{contradicts} for answer alignment. We provide keyword list, representative justifications, and additional scores in Appendix~\ref{app:qualan-findings}.
\vspace{-2mm}

\subsection{Individual Edits}
So far, we have focused on edits that do not change the final answer, which raises the question of whether models rely on pattern matching to recover the original answer. Individual edits, together with explicit tracking of the resulting answer, allow us to investigate this behavior. Since each record can have multiple edited instances, we use grouped accuracy to measure the effect of single edits, both with and without changes to the final answer. Specifically, we measure the difference in performance between edited records and the model’s original performance on the corresponding unedited records. For a fair comparison, we also apply single edits that do not change the final answer and compare model behavior across these two conditions. We assess statistical significance using a pairwise test.

\input{norm_logic/memorzation}
Table~\ref{tab:single_delta} reports the performance changes between edit level 0 and the grouped single-edit conditions.\footnote{We exclude AR-LSAT because only 22 records undergo label-changing edits which is too few for reliable estimation.} For deduction tasks with premise--hypothesis structures, such as FOLIO, performance consistently drops when the label changes, with large declines observed across high-performing models. Although the larger Gemma models generally achieve high accuracy on logical deduction, their performance decreases when labels are changed through small operator-level edits. This pattern may indicate that the models rely on patterns associated with the original questions and answers rather than robust reasoning. We explore this further by analyzing model prediction behavior.
\vspace{-1mm}

\paragraph{Prediction Behavior}
We measure how often a model falls back on the original answer instead of predicting the edited answer. Table~\ref{tab:memorisationrates} reports this behavior. We define the ideal case as one in which the model does not output the original answer after a label-changing edit. The alternative hypothesis is that the model outputs the original answer for more than 10\% of the edited records.

Except for Qwen, all models show some level of original-answer retention on FOLIO. Since Qwen previously showed a significant performance drop under label-changing edits, its behavior may be attributed to reasoning difficulty rather than original-answer retention. The Logical Deduction dataset also shows evidence of original-answer retention among lower-performing models; however, as model size increases, the retention rate decreases. Under a lower threshold of 5\%, all models except Qwen show signs of original-answer retention. 
\vspace{-1mm}

%% file: norm_logic/transition-small.tex
\begin{table}[t]
\centering
\small
\setlength{\tabcolsep}{3pt}
\resizebox{\columnwidth}{!}{
\begin{tabular}{
ll
S[table-format=2.2]
S[table-format=2.2]
S[table-format=2.2]
S[table-format=2.2]
S[table-format=2.2]
S[table-format=2.2]
}
\toprule
& \multirow{2}{*}{$k$}
& \multicolumn{3}{c}{\texttt{Gemma3-27B}}
& \multicolumn{3}{c}{\texttt{Qwen3-4B}} \\
\cmidrule(lr){3-5} \cmidrule(lr){6-8}
&
& \multicolumn{1}{c}{CoT}
& \multicolumn{1}{c}{Scaling}
& \multicolumn{1}{c}{SymbCoT}
& \multicolumn{1}{c}{CoT}
& \multicolumn{1}{c}{Scaling}
& \multicolumn{1}{c}{SymbCoT} \\
\midrule

\multirow{3}{*}{\rotatebox{90}{FOLIO}}
& 0 & 81.20 & 82.91 & 76.07 & 88.03 & 87.18 & 76.92  \\
& 1 & 82.05 & 85.47 & 78.63 & 87.18 & 88.03 & 76.92  \\
& 2 & 82.05 & 84.62 & 76.07 & 84.62 & 84.62 & 73.50  \\
\hline

\multirow{5}{*}{\rotatebox{90}{Deduction}}
& 0 & 84.33 & 93.00 & 86.00 & 96.00 & 95.33 & 74.00  \\
& 1 & 85.67 & 93.67 & 85.67 & 95.67 & 94.33 & 73.33  \\
& 2 & 84.67 & 94.33 & 86.00 & 95.67 & \textbf{92.33} & 73.67  \\
& 3 & 85.33 & 93.00 & 84.67 & 95.33 & \textbf{92.67} & {71.33}  \\
& 4 & 84.67 & 93.33 & 85.67 & 95.67 & 95.00 & 73.67  \\
\hline

\multirow{4}{*}{\rotatebox{90}{AR-LSAT}}
& 0 & 32.61 & 36.52 & 28.70 & 84.78 & 36.52 & 17.83  \\
& 1 & 33.04 & 37.39 & 30.43 & 83.04 & 35.65 & 19.13  \\
& 2 & 32.17 & \textbf{39.57} & 27.83 & 82.17 & 35.22 & 17.39  \\
& 3 & 34.35 & 36.96 & 28.70 & 83.48 & 36.09 & 16.52  \\
\hline
\end{tabular}}
\caption{Normalized Accuracy (\%) versus the number of label-preserving operator edits across datasets for Gemma 27B and Qwen 4B on valid records. Accuracy is computed over the corresponding total count, and \textbf{bold} indicates statistically significant performance changes, either gains or drops, relative to the base performance.
}
\vspace{-2mm}
\label{tab:backfilled_accuracy}
\end{table}

%% file: norm_logic/transition_table_perc.tex
\begin{table*}[t]
\centering
\small
\setlength{\tabcolsep}{3.5pt}
\begin{tabular}{
ll
c
S[table-format=2.2]
S[table-format=2.2]
S[table-format=2.2]
S[table-format=2.2]
S[table-format=2.2]
S[table-format=2.2]
S[table-format=2.2]
}
\toprule
&
{Method} &
{Transition} &
\multicolumn{1}{c}{\texttt{Gemma3-1B}} &
\multicolumn{1}{c}{\texttt{Gemma3-4B}} &
\multicolumn{1}{c}{\texttt{Gemma3-12B}} &
\multicolumn{1}{c}{\texttt{Gemma3-27B}} &
\multicolumn{1}{c}{\texttt{Llama3-8B}} &
\multicolumn{1}{c}{\texttt{Qwen3-4B}} &
\multicolumn{1}{c}{\texttt{Phi4Mini-4B}} \\
\midrule
\multirow{6}{*}{\rotatebox[origin=c]{90}{FOLIO}}
& \multirow{2}{*}{CoT}
& 0$\to$1  & \textbf{41.18} & 19.12 & 19.12 & 16.18 & \textbf{23.53} & 10.29 & \textbf{29.41} \\
& & 0$\to$2  & 23.08 & 19.23 & \textbf{34.62} & 26.92 & 26.92 & 23.08 & \textbf{46.15} \\
\addlinespace[1pt]
 \cdashline{2-10}
 \addlinespace[1pt]
& \multirow{2}{*}{Scaling}
& 0$\to$1  & 2.94 & 13.24 & 2.94 & 10.29 & 14.71 & 7.35 & 16.18 \\
& & 0$\to$2  & 3.85 & 15.38 & 30.77 & 23.08 & 19.23 & 11.54 & 26.92 \\
\addlinespace[1pt]
 \cdashline{2-10}
 \addlinespace[1pt]
& \multirow{2}{*}{SymbCoT}
& 0$\to$1  & \textbf{29.41} & \textbf{33.82} & \textbf{33.82} & \textbf{27.94} & 8.82 & 14.71 & 22.06 \\
& & 0$\to$2  & 30.77 & 23.08 & \textbf{46.15} & 23.08 & 15.38 & \textbf{46.15} & 19.23 \\
\hline
\multirow{12}{*}{\rotatebox[origin=c]{90}{Deduction}}
& \multirow{4}{*}{CoT}
& 0$\to$1  & \textbf{19.35} & \textbf{31.45} & 9.68 & 9.68 & \textbf{23.39} & 4.03 & \textbf{22.58} \\
& & 0$\to$2  & \textbf{23.36} & \textbf{33.64} & 14.95 & 12.15 & \textbf{26.17} & 2.80 & \textbf{28.97} \\
& & 0$\to$3  & 20.48 & \textbf{39.76} & 9.64 & 8.43 & \textbf{25.30} & 2.41 & 20.48 \\
& & 0$\to$4  & \textbf{28.00} & \textbf{30.00} & 8.00 & 6.00 & 22.00 & 2.00 & \textbf{26.00} \\
\addlinespace[1pt]
 \cdashline{2-10}
 \addlinespace[1pt]
& \multirow{4}{*}{Scaling}
& 0$\to$1  & 12.90 & 16.13 & 2.42 & 4.84 & 8.06 & 4.03 & 11.29 \\
& & 0$\to$2  & 12.15 & 18.69 & 4.67 & 5.61 & 3.74 & 10.28 & 8.41 \\
& & 0$\to$3  & 19.28 & 18.07 & 0.00 & 0.00 & 3.61 & 9.64 & 3.61 \\
& & 0$\to$4  & \textbf{28.00} & 22.00 & 0.00 & 2.00 & 8.00 & 6.00 & 6.00 \\
\addlinespace[1pt]
 \cdashline{2-10}
 \addlinespace[1pt]
& \multirow{4}{*}{SymbCoT}
& 0$\to$1  & 8.87 & 12.10 & 12.10 & 4.03 & \textbf{20.16} & 14.52 & 17.74 \\
& & 0$\to$2  & 5.61 & 17.76 & \textbf{20.56} & 7.48 & \textbf{20.56} & 19.63 & \textbf{26.17} \\
& & 0$\to$3  & 7.23 & \textbf{25.30} & 15.66 & 4.82 & \textbf{26.51} & 19.28 & \textbf{30.12} \\
& & 0$\to$4  & 10.00 & 22.00 & 20.00 & 2.00 & \textbf{26.00} & 18.00 & 24.00 \\
\hline
\multirow{9}{*}{\rotatebox[origin=c]{90}{LSAT}}
& \multirow{3}{*}{CoT}
& 0$\to$1  & \textbf{15.31} & \textbf{13.27} & \textbf{23.47} & \textbf{21.43} & \textbf{29.59} & \textbf{18.37} & \textbf{17.35} \\
& & 0$\to$2  & \textbf{23.21} & \textbf{14.29} & \textbf{23.21} & \textbf{23.21} & \textbf{19.64} & \textbf{21.43} & \textbf{26.79} \\
& & 0$\to$3  & \textbf{15.38} & 11.54 & \textbf{42.31} & \textbf{23.08} & \textbf{15.38} & 11.54 & \textbf{26.92} \\
\addlinespace[1pt]
 \cdashline{2-10}
 \addlinespace[1pt]
& \multirow{3}{*}{Scaling}
& 0$\to$1  & 8.16 & 8.16 & \textbf{10.20} & \textbf{14.29} & \textbf{10.20} & \textbf{12.24} & \textbf{13.27} \\
& & 0$\to$2  & 5.36 & 5.36 & 7.14 & \textbf{16.07} & \textbf{12.50} & \textbf{16.07} & \textbf{25.00} \\
& & 0$\to$3  & 11.54 & 0.00 & \textbf{23.08} & 3.85 & 7.69 & 11.54 & \textbf{15.38} \\
\addlinespace[1pt]
 \cdashline{2-10}
 \addlinespace[1pt]
& \multirow{3}{*}{SymbCoT}
& 0$\to$1  & \textbf{15.46} & \textbf{15.31} & \textbf{32.65} & \textbf{32.65} & \textbf{10.20} & 9.18 & \textbf{22.45} \\
& & 0$\to$2  & \textbf{25.00} & \textbf{14.29} & \textbf{37.50} & \textbf{28.57} & \textbf{12.50} & 8.93 & \textbf{19.64} \\
& & 0$\to$3  & \textbf{15.38} & 7.69 & \textbf{23.08} & \textbf{38.46} & 0.00 & 11.54 & \textbf{15.38} \\
\hline
\end{tabular}
\caption{Flip Rate: Total Percentage of records that transition from correct to incorrect and from incorrect to correct as logic-forms change from edit 0 to later edits. Higher values indicate greater instability across edits. Values in \textbf{bold} indicate significance of transitions across all models. 
}
\vspace{-2mm}
\label{tab:moved-pct-zero-to-k}
\end{table*}

%% file: norm_logic/qual_analysis.tex
\begin{table}[t]
\small
\centering
\setlength{\tabcolsep}{11pt}
\begin{tabular}{
S[table-format=1.0]
S[table-format=2.2]
S[table-format=2.2]
S[table-format=2.2]
S[table-format=2.2]
}
\toprule
{Edits} & {PF} & {IV} & {AC} & {AL} \\
\midrule
0 & 83.33 & 75.00 & 91.67 & 87.50 \\
1 & 79.17 & 70.83 & 79.17 & 75.00 \\
2 & 70.83 & 66.67 & 75.00 & 83.33 \\
3 & 75.00 & 58.33 & 70.83 & 62.50 \\
\bottomrule
\end{tabular}
\caption{Qualitative scores (\%) by the number of label-preserving edits for the Qwen3-4B model on the AR-LSAT dataset. The results show a decline in reasoning metrics as the number of edits increases.}
\vspace{-3mm}
\label{tab:scores_by_edits}
\end{table}

%% file: norm_logic/singleedits.tex
\begin{table*}[t]
\centering
\small
\begin{tabular}{
ll
S[table-format=-2.2]
S[table-format=-2.2]
S[table-format=-2.2]
S[table-format=-2.2]
S[table-format=-2.2]
S[table-format=-2.2]
S[table-format=-2.2]
}
\toprule
&
Condition &
\multicolumn{1}{c}{\texttt{Gemma3-1B}} &
\multicolumn{1}{c}{\texttt{Gemma3-4B}} &
\multicolumn{1}{c}{\texttt{Gemma3-12B}} &
\multicolumn{1}{c}{\texttt{Gemma3-27B}} &
\multicolumn{1}{c}{\texttt{Llama3-8B}} &
\multicolumn{1}{c}{\texttt{Qwen3-4B}} &
\multicolumn{1}{c}{\texttt{Phi4-4B}} \\
\midrule

\multirow{2}{*}{\textbf{FOLIO}} & Original
& -5.59
& 2.38
& 5.93
& 3.55
& 5.29
& -0.12
& -0.47 \\
& Altered
& {\textbf{-27.33}}
& {\textbf{-29.33}}
& -16.00
& {\textbf{-23.33}}
& -21.33
& {\textbf{-22.67}}
& -4.67 \\
\midrule

\multirow{2}{*}{\textbf{Deduction}}& Original
& -4.00
& -2.91
& 3.40
& 2.79
& -4.05
& -0.68
& 3.13 \\
& Altered
& 2.87
& -0.32
& {\textbf{-7.70}}
& {\textbf{-12.95}}
& -8.65
& -0.32
& -8.23 \\
\bottomrule
\end{tabular}
\caption{Difference in model accuracy(\%) under operator mutation with and without alteration. \textbf{Bold} values indicate statistically significant differences. Negative values represent drops in accuracy, whereas positive values represent gains. The original condition refers to the existing label, while the altered condition refers to the tool-edited label.}
\label{tab:single_delta}
\vspace{-2mm}
\end{table*}

%% file: norm_logic/memorzation.tex
\begin{table}[t]
\centering
\small
\setlength{\tabcolsep}{11pt}
\begin{tabular}{l S[table-format=2.2] S[table-format=2.2]}
\toprule
{Model} & {FOLIO (\%)} & {Deduction (\%)} \\
\midrule
\texttt{Gemma3-1B}   & {\textbf{36.23}} & {\textbf{24.00}} \\
\texttt{Gemma3-4B}   & {\textbf{30.43}} & 13.14 \\
\texttt{Gemma3-12B}  & {\textbf{28.99}} & 17.14 \\
\texttt{Gemma3-27B}  & {\textbf{27.54}} & 13.14 \\
\texttt{Llama3-8B}   & {\textbf{42.03}} & 16.00 \\
\texttt{Qwen3-4B}    & 15.94            & 3.43 \\
\texttt{Phi4Mini-4B} & {\textbf{23.19}} & {\textbf{18.86}} \\
\bottomrule
\end{tabular}
\caption{Original-answer retention rates by model and dataset. Bold values indicate statistically significant retention rate at $p < 0.05$ using a 10\% threshold.}
\vspace{-3mm}
\label{tab:memorisationrates}
\end{table}

%% file: norm_logic/Conclusion.tex
Understanding how LLM reasoning varies with structural changes is key to assessing the extent to which models follow logical structure. In this work, we present a tool-driven framework that generates controlled, label-preserving edits by operating on symbolic representations of logical reasoning problems. This enables precise operator-level interventions while preserving the original deductive outcome. Using this framework, we analyze LLM behavior and show that models reason inconsistently under these edits. These findings suggest that current LLMs do not reliably ground their reasoning in symbolic structure, highlighting the need to benchmark these models for structural reasoning and consider evaluation metrics beyond standard accuracy.

Overall, our evaluation provides a controlled way to stress-test LLMs and measure their robustness to structural variations. Future work could use stability under such variations as a training or evaluation signal, helping to reduce reliance on pattern matching. The proposed template-generation framework may also support the construction of more diverse benchmarks through symbolic editing, exposing models to a broader range of reasoning patterns during training and evaluation.

%% file: norm_logic/Appendix.tex
\section{Symbolic Translations}
\label{app:translations}

We extract symbolic translations based on the dataset and the availability of existing translations. For Logical Deduction and AR-LSAT, symbolic representations are generated following the two-shot prompting setup of \citet{pan2023logic}, adapted to a structured JSON format. This design allows us to enforce formatting constraints and guide the model toward producing well-formed outputs. Unless otherwise specified, all symbolic representations are generated using \texttt{Gemini 2.5 Flash} with a temperature of (0.1) under batched inference. The final dataset statistics are summarized in Table~\ref{tab:dataset_statistics}.

\begin{itemize}[leftmargin=*,nosep]
\item \textbf{FOLIO:} FOLIO provides ground-truth symbolic representations that do not always follow the tool format. We first run these through Prover9 and manually edit a set of formulations to correct annotation errors. These errors are identified by inspecting tool feedback. The corrected symbolic forms are retained for approximately 10\% of the records.

\item \textbf{Logical Deduction:} Logical Deduction is relatively easy to translate, with the model achieving a 99\% success rate on the first generation pass. We run the model twice with a temperature of (0.1) to obtain all 300 translations suitable for execution using a Python-based CSP tool.

\item \textbf{AR-LSAT:} AR-LSAT is more challenging to translate into a form suitable for Z3. We initially use a temperature of (0.1) and then increase decoding randomness. We find that test-time scaling with a temperature of (0.7), an incremental scaling factor of 4, and up to 8 samples yields a larger number of valid translations. This incremental scaling strategy reduces inference cost by 12.7\% compared to directly generating 8 samples for every record. Using \texttt{Gemini 2.5 Flash}, we extract translations for approximately 88\% of the records. For the remaining records, we use \texttt{Gemini 3 Flash Preview} with diverse decoding strategies by varying temperature, top-(k), and top-(p) values. We note that self-correction does not improve translation quality over this inference-only approach with diverse decoding strategies. Overall, we obtain 224 translated records. Among these, we identify 6 records with incorrect final labels, which the authors manually correct by solving the questions and verifying the answers with multiple LLMs. These corrections are reported in Table~\ref{tab:arlsat_label_corrections} and are used as the ground truth for all results.
\end{itemize}

\begin{table}[t]
\centering
\small
\setlength{\tabcolsep}{11pt}
\begin{tabular}{l l r }
\hline
\textbf{Dataset} & \textbf{Tool} & \textbf{Valid} \\
\hline
FOLIO & Prover9 & 117   \\
Deduction & Python & 300  \\
AR-LSAT & Z3 & 224 \\
\hline
\end{tabular}
\caption{Statistics of datasets showing the number of records with valid translations 
}
\label{tab:dataset_statistics}
\end{table}

\begin{table}[t]
\centering
\small
\begin{tabular}{lll}
\toprule
\textbf{ID} & \textbf{Given} & \textbf{Corrected} \\
\midrule
\texttt{ar\_lsat\_201612\_3-G\_1\_3}  & B & D \\
\texttt{ar\_lsat\_201612\_3-G\_1\_5}  & E & B \\
\texttt{ar\_lsat\_201612\_3-G\_2\_6}  & C & D \\
\texttt{ar\_lsat\_201612\_3-G\_2\_8}  & A & B \\
\texttt{ar\_lsat\_201612\_3-G\_3\_12} & C & A \\
\texttt{ar\_lsat\_201612\_3-G\_3\_17} & D & E \\
\bottomrule
\end{tabular}
\caption{Corrections to AR-LSAT answer labels. The \textit{Given} column shows the original label, and the \textit{Corrected} column shows the revised label.}
\label{tab:arlsat_label_corrections}
\end{table}

\section{Edit Stats}
\label{app:datastat}
After applying cumulative and single edits, we perform back-translation of the logical forms using the corresponding natural-language text provided with each logical form. In some cases, the natural-language sentence associated with a logical form does not appear as a standalone sentence in the original question. This occurs, for example, in AR-LSAT, where a statement such as ``there are four objects'' may be embedded within a longer sentence rather than appearing independently. In other cases, the LLM performing back-translation fails to generate a valid replacement, leaving the final question unchanged.

After eliminating cases that do not support the required back-translate-and-replace procedure, fewer records remain. For cumulative edits, if a translation is unavailable at edit level ($k$), all subsequent edits are discarded, and the record is retained only up to edit level ($k-1$). The final counts of edited records are reported in Table~\ref{tab:datastat-edits}.

\begin{table}[t]
\centering
\small
\setlength{\tabcolsep}{7pt}
\begin{tabular}{l  r r r}
\hline
\textbf{Dataset} & \textbf{Cuml}& \textbf{Ind-original} & \textbf{Ind-altered} \\
\hline
FOLIO & 94 &68& 50 \\
Deduction & 364 &123& 157 \\
AR-LSAT & 180 &-&- \\
\hline
\end{tabular}
\caption{Total number of edits available for each dataset under two edit settings. Individual edits include both label-preserving edits and label-altering edits. AR-LSAT does not generate enough records to support a meaningful individual-edit analysis.
}
\label{tab:datastat-edits}
\end{table}

\section{Back-Translations}
\label{app:bt}
We apply back-translation at the sentence level by passing the original natural-language sentence as the incorrect statement and prompting the model to generate a corrected natural-language sentence without deviating from the semantic structure of the question. Figures~\ref{fig:logic-correction-icl}, \ref{fig:logic-correction-icl-deduction}, and \ref{fig:logic-correction-icl-folio} provide examples of the prompts used for AR-LSAT, Logical Deduction, and FOLIO, respectively.

The Logical Deduction dataset involves simple alterations; however, the semantic variation induced by these changes is not easily programmable. For example, changing `less than'' to `greater than'' may correspond to different interpretations, such as older vs.\ newer or left vs.\ right, making it difficult to generalize such transformations.

\subsection*{Human Evaluation}
Back-translation is performed at the sentence level, where the LLM is provided with the formal statement and the original sentence, ensuring that the generated output follows a similar structure. To ensure that the back-translations maintain the expected fidelity, we sample outputs from each dataset, covering natural-language sentences of varying lengths (measured by the number of words in a sentence), and use the interactive form (\href{https://github.com/RamyaKeerthy/LogiEval-BT\_form.git}{RamyaKeerthy/LogiEval-BT\_form}) to evaluate their accuracy. Figure~\ref{fig:backtranslation_instructions} shows the instructions provided to the annotators at the beginning of the task. 

We conducted a human evaluation with three annotators, each holding a Master's degree or higher and having relevant experience in this domain. We sampled 30 examples from each dataset (90 in total), covering both short and long sentences and logical forms. Using detailed annotation guidelines, the annotators achieved 94.5\% inter-annotator agreement, measured using Gwet's AC1. Table~\ref{tab:human_evaluation} shows validity on majority annotations. The only majority-rejected example occurred in FOLIO (29/30 accepted). Upon closer examination of the FOLIO disagreements, we identified one major back-translation error in which the edited logical operator was not correctly reflected as the initial formal representation had incorrectly captured the semantics. Evaluating this case across four Gemma models showed no change in model predictions, indicating a negligible practical impact from this error. These results provide additional confidence in both the back-translation process and our prompting strategy.

\begin{figure*}[t]
\centering

\begin{tcolorbox}[
    colback=yellow!8,
    colframe=blue!45!black,
    boxrule=0.6pt,
    arc=2mm,
    left=3mm,
    right=3mm,
    top=2mm,
    bottom=2mm,
    title=\textbf{Human Evaluation of Back-Translation- Task Instructions},
    fonttitle=\small,
]

\small

\textbf{You are given:}
\begin{itemize}
    \setlength{\itemsep}{1pt}
    \setlength{\parskip}{0pt}
    \item A reference formal-logic expression.
    \item A reference natural-language translation of the reference formal logic.
    \item A target formal-logic expression.
    \item A final natural-language statement.
\end{itemize}

The reference natural-language translation shows how the reference formal logic is expressed in natural language. Use it to understand both the original meaning and the expected sentence structure.

The reference formal logic and the target formal logic differ by exactly one highlighted logical change. This highlighted change alters the meaning of the sentence and should be reflected in the final natural-language statement.

\medskip
\textbf{Your task} is to determine whether the final natural-language statement correctly and completely matches the target formal logic while maintaining the structure of the reference natural-language translation.

\medskip
\textbf{Answer YES only if:}
\begin{itemize}
    \setlength{\itemsep}{1pt}
    \setlength{\parskip}{0pt}
    \item The final natural-language statement accurately expresses the target formal logic.
    \item The highlighted logical change is correctly reflected in the final natural-language statement.
    \item The sentence structure closely follows the reference natural-language translation, except where a structural change is required to express the logical change.
    \item The content remains the same as the reference natural-language translation, apart from modifications required by the highlighted logical change.
\end{itemize}

\textbf{Answer NO if:}
\begin{itemize}
    \setlength{\itemsep}{1pt}
    \setlength{\parskip}{0pt}
    \item The highlighted logical change is not reflected correctly.
    \item The final natural-language statement changes the meaning of the target formal logic.
    \item The final natural-language statement omits, adds, or alters information beyond what is required by the highlighted logical change.
    \item The sentence structure is substantially different from the reference natural-language translation.
    \item The final natural-language statement is a paraphrase of the reference natural-language translation rather than a direct adaptation reflecting the logical change.
\end{itemize}

\medskip
Focus on both semantic accuracy and structural consistency with the reference natural-language translation. Ignore differences only when they are necessary to correctly express the highlighted logical change.
\end{tcolorbox}

\caption{Instructions used to evaluate the natural-language back-translations to the target formal-logic representation.}
\label{fig:backtranslation_instructions}
\end{figure*}

\begin{table}[t]
\centering
\small
\setlength{\tabcolsep}{5pt}
\begin{tabular}{lrrr}
\toprule
\textbf{Dataset} & \textbf{Total} & \textbf{Valid} & 
\textbf{Valid (\%)}  \\
\midrule
AR-LSAT    & 30 & 30 & 100.0  \\
FOLIO      & 30 & 29 & 96.7   \\
Deduction  & 30 & 30 & 100.0  \\
\midrule
\textbf{Overall} & \textbf{90} & \textbf{89} & \textbf{98.9}  \\
\bottomrule
\end{tabular}
\caption{Summary of human evaluation results across datasets. }
\label{tab:human_evaluation}
\end{table}

\section{Reasoning Prompts}
\label{app:reasonprompt}
We use the following system prompt for CoT and scaled CoT to obtain structured outputs: `You are a reasoning assistant that reasons step by step, and puts your final answer within \textbackslash boxed\{\}.' The user instruction explicitly prevents the use of external knowledge: `Solve the given logical deductive reasoning problem using only the information provided in the question. Do not use or rely on any external knowledge.'

For Symbolic CoT, we use the prompts from the original paper to perform the three stages of inference. The only modification is that we ask the model to generate JSON output, which allows us to extract the required inference results more easily. Apart from this formatting change, the examples and instructions remain the same.

The configurations for the models used to generate inferences are detailed in Appendix~\ref{app:configs}.

\section{Complete Results}
\label{app:cuml}
\subsection*{Cumulative Performance}

We present the performance of other open-source models on cumulative edits in Table~\ref{tab:backfilled_accuracy_combined}. Performance is measured at each edit level with respect to the base question. If an edit is not available for a given input, we fall back to the base question, ensuring that the total number of evaluated records remains constant across edit levels. The number of records affected at each edit level is reported as \textit{Count} in the table. The results show minimal impact on performance across edits. However, models such as Gemma 1B and Phi4mini show a drop in performance at higher edit levels under chain-of-thought reasoning. The stable accuracy values do not capture the full extent of the variation occurring across edits, motivating us to further examine prediction flip rates.

To extend this analysis to reasoning models, we also evaluate Gemini 2.5 Flash, which uses default thinking mode. The results are presented in Table~\ref{tab:gemini_flip_significance}. Its behavior closely matches that of Gemma-27B while exhibiting greater overall stability than the smaller open models. This observation highlights that reasoning consistency depends on the dataset type: challenging datasets such as AR-LSAT continue to exhibit prediction flips even for larger reasoning models, whereas Logical Deduction remains highly stable.

\begin{table}[t]
\centering
\small
\begin{tabular}{lccc}
\toprule
\textbf{Dataset} & \textbf{Edit Level} & \textbf{Flip Rate} & \textbf{$p$-value} \\
\midrule

\multirow{2}{*}{FOLIO}
    & 0$\to$1 & 11.8\% & 0.3698 \\
    & 0$\to$2 & 19.2\% & 0.1118 \\
\midrule

\multirow{3}{*}{Deduction}
    & 0$\to$1 & 0.8\% & 1.0000 \\
    & 0$\to$2 & 0.0\% & 1.0000 \\
    & 0$\to$3 & 0.0\% & 1.0000 \\

\midrule
\multirow{3}{*}{AR-LSAT}
    & 0$\to$1 & 11.2\% & 0.3905 \\
    & 0$\to$2 & 16.1\% & 0.1030 \\
    & \textbf{0--3} & \textbf{23.1\%} & \textbf{0.0399} \\

\bottomrule
\end{tabular}

\caption{Flip rates for \texttt{Gemini 2.5 Flash} under chain-of-thought reasoning between the baseline (0) and cumulative edit levels. Statistically significant results ($p < 0.05$) are shown in \textbf{bold}.}
\label{tab:gemini_flip_significance}
\end{table}

\input{norm_logic/transition_edit0}

\subsection*{Single Edits}
We present the accuracy results for all single edits in Table~\ref{tab:singleedits-all}. Accuracy differences are computed as the difference between base accuracy and edited-record accuracy. Labels are either preserved or changed using the tool wrapper. The number of base records represents the number of unique records with valid edits among the total available translations.

Label-changing edits are fewer than label-preserving edits, since extracting a completely new label from a given question often requires substantially different reasoning. This new reasoning requirement may also explain the drastic drop in accuracy under altered labels. Overall, this provides a useful way to evaluate existing datasets that have reached near-saturated performance and to test whether models are still reasoning correctly under controlled structural changes.

\input{norm_logic/qualitative-full-results}
\input{norm_logic/singleedits_full}

\section{Variation in proof}
\label{app:variationinproof}
Prover9 can output proof length (the total number of input and derived clauses used in the proof) that can work as a proxy for proof variation. This metric is only available for examples with deterministic labels (True/False). Among these, only 15 records in total showed a change in proof length for cumulative edits. We then examined whether change in proof-length correlated with changes in model predictions. Across the Gemma models, only Gemma-4B showed a noticeable trend, with larger proof-length changes more likely to induce prediction flips. Strong conclusions are avoided due to the limited available proofs.

Where proof statistics are unavailable using a tool, we measured changes in reasoning length using a proprietary model (Gemini 2.5 Flash). Considering examples whose reasoning length changed a lot, for instance, records with token difference greater than 250 (includes drop and gain in token size), we found that no two models consistently flipped on the same examples. For both LogicalDeduction and AR-LSAT, at least one model remained stable in its prediction despite changes in reasoning length.

\section{Qualitative Analysis- Prompt}
\label{app:qualan}
In the qualitative analysis, we examine model reasoning patterns in a generated CoT setting. For logical deduction tasks, the reasoning process typically proceeds in three steps: first, the model identifies the premises provided in the question; second, it derives intermediate inferences from these premises when necessary; and third, it checks whether one of the answer options is supported by the reasoning. We also examine the model’s final answer claim, especially the answer placed in \texttt{\textbackslash boxed}, as requested in the prompt.

Based on this observed reasoning process, we divide the qualitative analysis into four components. We define the coding criteria as Premise Fidelity (PF), Inference Validity (IV), Answer Completion (AC), and Answer Alignment (AL). Each code is designed to measure a specific aspect of the reasoning process. These criteria are described in more detail in the evaluation prompt shown in Figure~\ref{fig:qualitative-code-prompt}. The LLM evaluator receives both the original question and the generated reasoning path as input.

This analysis depends on the verification capability of the evaluator LLM. Therefore, the results should be interpreted as estimated judgments rather than conclusive findings, and further study is needed to validate these observations.

\section{Qualitative Analysis- Findings}
\label{app:qualan-findings}
Table~\ref{tab:appendix-full-qualitative} presents the full set of LLM-evaluated qualitative results for records with the maximum number of edits in each dataset. Smaller models generally show low premise fidelity and inference validity. Models such as Gemma3-1B also exhibit very low answer completeness, indicating missing or incomplete final answers, as well as low answer alignment, which is correlated with answer completion. Cases with high answer alignment but low answer completeness suggest that the model may rely on incorrect assumptions to arrive at the final answer. The opposite pattern suggests that the model changes its final decision despite partially complete reasoning. All models except Qwen3-4B show reasoning issues on AR-LSAT, which is also reflected in their accuracy.

We also examine individual tokens appearing in the justifications for each code, with an emphasis on negative tokens. We generate justifications only for criteria assigned a score of (0), enabling targeted negative-token analysis. Table~\ref{tab:top20_justification_tokens} reports the top 20 tokens for each code across all models and edits. We highlight tokens that provide additional insight into common reasoning failures. Representative justifications based on selected keywords are provided below.

\begin{itemize}
\item \textbf{distorts:} The reasoning invents a crucial premise and \textbf{distorts} another. It states that `all of Kelly Wearstler's designs have interesting geometries,' which is not supported by the original text. The original premises state that these designs are `evocative' and `dreamy.' It also \textbf{distorts} another premise by changing the antecedent of the conditional statement. Specifically, it rephrases `If a design by Max that he adores has interesting geometries, then the design is a brutalist building and evocative' as `if a design by Max is a design style that is Zaha Hadid's or Kelly Wearstler's, then it is a brutalist building'.

\item \textbf{misinterprets:} The reasoning \textbf{misinterprets} several key premises. For option A (Louis), it states: 'If Louis is assigned to 1922, then Tiffany cannot be assigned to 1923.' This distorts condition C1 (Louis=1923 XOR Tiffany=1923), which actually implies that if Louis is NOT in 1923 (e.g., in 1922), then Tiffany *must* be in 1923. For option C (Onyx), it \textbf{misinterprets} condition C4, incorrectly deducing that 'Ryan must be assigned to 1923 or 1922' when Onyx is in 1922. For option D (Ryan), it \textbf{misinterprets} condition C3 (Tiffany\_assigned -> Ryan\_assigned) by stating 'If Ryan is assigned to 1922, then Tiffany must be assigned,' which is the converse and not necessarily true.

\item \textbf{invents:} The reasoning \textbf{invents} a premise by stating '1. Bird Order: The birds are arranged in a fixed order: blue jay, owl, falcon, hawk, raven, crow, hummingbird.' This is a list of the birds' names, not their fixed arrangement. It then \textbf{invents} a specific bird order ('owl, blue jay, raven, falcon, hawk, hummingbird, crow') which contradicts the given premise that 'The falcon is the second from the left' (in the invented order, the falcon is fourth).

\item \textbf{leap:} The reasoning contains multiple invalid logical steps and unsupported \textbf{leaps}. In Scenario 1, the deduction that \(P\) must be performed before \(O\) is unsupported, and the subsequent deductions about the positions of \(P\) and \(S\) rely on this invalid leap. More critically, the reasoning misses a direct contradiction: if \(R=1\) and \(F=2\), then \(T\) must be 1 by condition 1, which places both \(T\) and \(R\) at position 1. In Scenario 2, the inference that \(T\) cannot be performed immediately before \(F\) is a direct logical error based on condition 1. The inference that \(O\) must be first is also invalid given the constraint \(F \_ \_ O\). The statement that the remaining compositions \(L\), \(H\), and \(R\) must be performed fourth, fifth, and sixth directly contradicts the scenario’s initial premise that \(R\) is performed last.

\item \textbf{skips:} The reasoning \textbf{skips} the crucial intermediate steps needed to systematically combine the relative-position constraints and derive the fixed order of the birds. It presents an assumed order without derivation, identifies a contradiction, reaches a premature conclusion, and then makes an unsupported jump to the final correct conclusion about the leftmost bird without showing the valid intermediate steps.

\item \textbf{contradicts:} The final answer, ``D) The robin is the rightmost,'' \textbf{contradicts} the model's own derived sequence: Quail, Hummingbird, Robin, Raven, Blue Jay, Falcon, in which Falcon is the rightmost. It also \textbf{contradicts} the model's own evaluation of option D, which it marks as inconsistent with the derived order.

\end{itemize}

These tokens show certain behaviors of large language models that needs to be studied in detail. This qualitative analysis is a starting step and provides valuable insights into the internal tendencies of LLMs and supports the development of more rigorous evaluation methods.

\begin{table*}[ht]
\centering
\small
\begin{tabular}{lllll}
\toprule
Premise Fidelity & Inference Validity & Completeness & Answer Alignment & Overall Reasoning \\
\midrule
reasoning & reasoning & reasoning & reasoning & reasoning \\
premise & \cellcolor{yellow!30}invalid & \cellcolor{yellow!30}fails & answer & answer \\
assigned & option & option & final & final \\
condition & assigned & steps & directly & logical \\
model & logical & constraints & option & \cellcolor{yellow!30}invalid \\
states & states & does & conclusion & premise \\
\cellcolor{yellow!30}distorts & model & \cellcolor{yellow!30}incomplete & supported & evaluation \\
constraint & \cellcolor{yellow!30}unsupported & intermediate & concludes & premises \\
\cellcolor{yellow!30}misinterprets & premise & fully & model & completeness \\
position & position & systematically & true & inference \\
original & \cellcolor{yellow!30}incorrectly & options & \cellcolor{yellow!30}flawed & dimensions \\
second & condition & valid & follows & significant \\
\cellcolor{yellow!30}invents & inference & conditions & based & \cellcolor{yellow!30}issues \\
option & steps & necessary & does & fidelity \\
performed & constraint & constraint & \cellcolor{yellow!30}contradicts & validity \\
premises & makes & correctly & options & \cellcolor{yellow!30}suffers \\
recorded & second & explore & states & \cellcolor{yellow!30}unsupported \\
given & \cellcolor{yellow!30}leap & complete & possible & fundamental \\
\cellcolor{yellow!30}incorrectly & does & \cellcolor{yellow!30}skips & correct & leading \\
stating & contains & condition & provided & inferences \\
\bottomrule
\end{tabular}
\caption{Top 20 Tokens Across Justification Categories with negative tokens highlighted. }
\label{tab:top20_justification_tokens}
\end{table*}

\section{LogicEdit Library}
\label{app:logicedit}
We introduce \emph{LogicEdit}, a Python library that uses symbolic reasoning solvers as wrappers to generate controlled operator edits for logical reasoning tasks. The library accepts a formal representation of a logical reasoning instance and produces edited versions according to user-specified requirements. It supports both single-edit generation, where each variation contains exactly one logical edit, and cumulative-edit generation, where successive edits are applied incrementally to increase the degree of structural variation. Users can additionally specify whether the final label should be preserved or altered. The edited symbolic representations are translated back into natural language to produce new benchmark instances.

\begin{figure*}[ht]
    \centering
    \includegraphics[width=\textwidth]{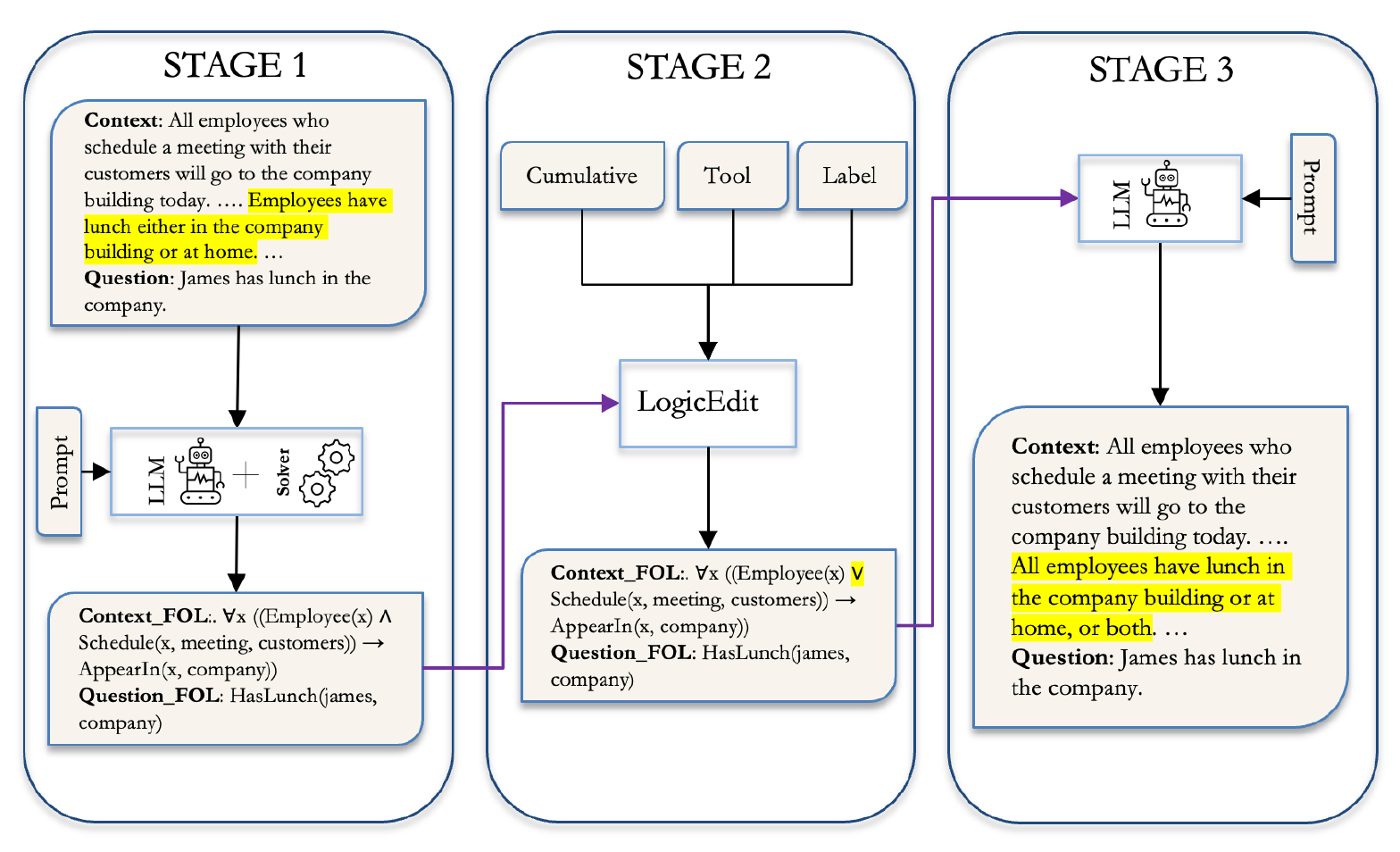}
    \caption{Pipeline integrating the LogicEdit library to generate alternative benchmark variants for testing the ability of LLMs to reason robustly.
    }
    \label{fig:logicedit}
\end{figure*}

Since LogicEdit requires formal translation before editing and back-translation after editing, we provide a complete pipeline that seamlessly integrates the library and offers an alternative option for users to generate logically verified benchmark variations suitable for stress-testing LLMs. This pipeline is included in the codebase associated with this work. Figure~\ref{fig:logicedit} outlines the complete pipeline, allowing the integration of the library.

\section{Models and Configurations}
\label{app:configs}
We use seven open-source models obtained from Hugging Face repositories. The Gemma models are run through the API for Chain-of-Thought and Symbolic CoT inference. Since the API does not support inference-time scaling, we run the scaling experiments on A100 GPUs. For the remaining models, we use A100 GPUs with 100GB memory and run inference with temperature (=0.1), unless otherwise specified.

For inference-time scaling, the number of samples for Gemma 27B is limited to 8 due to memory and resource constraints. For all other models, we generate 32 samples at temperature (=0.7). For CoT prompting, we use a maximum generation length of 10,000 tokens. For the other two prompting strategies, we limit generation to 2,000 tokens to manage memory usage. Any performance distortion, such as that observed for Qwen, may be due to the smaller output-token limit, which can lead to incomplete inferences on datasets such as AR-LSAT, where reasoning is substantially more complex.

For symbolic data generation, back-translation, and qualitative-analysis generation, we use \texttt{Gemini 2.5 Flash}, unless otherwise specified.

We use only open-source models that are publicly available through Hugging Face repositories. All datasets used in this work are publicly available and do not contain sensitive information.

\input{norm_logic/backtranslation-prompts}

\input{norm_logic/qualitative-prompt}

%% file: norm_logic/transition_edit0.tex
\begin{table*}[t]
\centering
\small
\scriptsize
\setlength{\tabcolsep}{2.5pt}
\renewcommand{\arraystretch}{1.08}

\begin{tabular}{ll|ccc|ccc|ccc|ccc|ccc|c}
\toprule
 & Edits
& \multicolumn{3}{c|}{Gemma3 1B}
& \multicolumn{3}{c|}{Gemma3 4B}
& \multicolumn{3}{c|}{Gemma3 12B}
& \multicolumn{3}{c|}{Llama3 8B}
& \multicolumn{3}{c|}{Phi4mini}
& Count \\
&
& CoT & Scaling & SymbCoT
& CoT & Scaling & SymbCoT
& CoT & Scaling & SymbCoT
& CoT & Scaling & SymbCoT
& CoT & Scaling & SymbCoT \\
\midrule

\multirow{3}{*}{\rotatebox{90}{FOLIO}}
& 0 & 47.01 & 41.03 & 35.90 & 71.79 & 74.36 & 51.28 & 80.34 & 89.74 & 71.79 & 58.12 & 70.94 & 44.44 & 68.38 & {72.65} & 24.79 & 117 \\
& 1 & 43.59 & 41.03 & 41.03 & 74.36 & 75.21 & 55.56 & 79.49 & 89.74 & 74.36 & 59.83 & 70.94 & 42.74 & 68.38 & {71.79} & 22.22 & 68 \\
& 2 & 47.01 & 41.88 & 39.32 & 69.23 & 74.36 & 52.99 & 81.20 & 89.74 & 70.09 & 60.68 & 70.09 & 44.44 & 64.96 & \textbf{66.67} & 25.64 & 26 \\

\midrule

\multirow{6}{*}{\rotatebox{90}{Logical Deduction}}
& 0 & {21.67} & 26.33 & 6.33 & 37.00 & 39.33 & 50.00 & 82.00 & 91.33 & 76.00 & 36.33 & 39.33 & 42.00 & 48.00 & {58.67} & 38.00 & 300 \\
& 1 & {20.67} & 26.33 & 8.00 & 35.33 & 39.33 & 49.00 & 81.33 & 91.67 & 77.00 & 35.00 & 39.00 & 44.00 & 48.67 & {59.67} & 37.33 & 124 \\
& 2 & {19.33} & 28.00 & 7.67 & 36.33 & 39.33 & 52.33 & 80.00 & 91.00 & 78.00 & 36.00 & 40.00 & 44.67 & 47.67 & {61.67} & 33.00 & 107 \\
& 3 & {21.67} & 27.67 & 7.67 & 35.33 & 40.33 & 52.67 & 83.33 & 91.33 & 77.67 & 35.67 & 39.33 & 44.00 & 48.33 & {61.33} & 35.67 & 83 \\
& 4 & {21.67} & 27.00 & 6.67 & 37.33 & 39.67 & 50.67 & 83.33 & 91.33 & 78.00 & 35.67 & 39.33 & 44.00 & 47.67 & {61.67} & 38.00 & 50 \\
& 5 & \textbf{24.33} & 28.67 & 6.67 & 36.67 & 38.67 & 50.67 & 82.67 & 91.33 & 77.67 & 36.00 & 39.33 & 41.67 & 49.00 & \textbf{62.67} & 39.33 & 41 \\

\midrule

\multirow{4}{*}{\rotatebox{90}{AR-LSAT}}
& 0 & 22.61 & 20.00 & 12.17 & 15.22 & 17.83 & 13.91 & 23.91 & 32.61 & 26.09 & 17.83 & 20.87 & 13.91 & 14.35 & 19.13 & 16.09 & 230 \\
& 1 & 21.30 & 22.61 & 10.87 & 14.78 & 18.70 & 14.35 & 23.48 & 33.48 & 22.61 & 15.65 & 20.00 & 13.04 & 14.35 & 20.43 & 16.52 & 98 \\
& 2 & 23.04 & 20.43 & 11.30 & 16.09 & 18.26 & 13.91 & 23.48 & 33.48 & 23.04 & 17.83 & 21.30 & 13.91 & 15.22 & 21.30 & 16.52 & 56 \\
& 3 & 22.61 & 21.30 & 12.17 & 15.65 & 17.83 & 14.35 & 21.74 & 33.48 & 25.22 & 16.96 & 20.87 & 13.91 & 14.78 & 19.57 & 16.96 & 26 \\

\bottomrule
\end{tabular}

\caption{Normalized accuracy (\%) versus the number of label-preserving operator edits across datasets and models. We report results for Gemma 1B, Gemma 4B, Gemma 12B, Llama 8B, and Phi4mini. \textbf{Bold} values indicate statistically significant records, where a larger drop in accuracy is captured. Count denotes the number of records corresponding to each edit level.}
\label{tab:backfilled_accuracy_combined}
\end{table*}

%% file: norm_logic/qualitative-full-results.tex
\begin{table}[ht]
\centering
\tiny
\setlength{\tabcolsep}{4pt}
\begin{tabular}{llrrrrrrrr}
\toprule
 & Dataset & Edits & PF & IV & AC & AL & Overall & Correct & Count \\
\midrule
\multirow{6}{*}{\rotatebox{90}{Gemma3-1B}}
& \multirow{2}{*}{AR-LSAT} & 0 & 3 & 0 & 0 & 1 & 1.00 & 1 & 24 \\
& & 3 & 0 & 0 & 0 & 2 & 0.50 & 2 & 24 \\
& \multirow{2}{*}{FOLIO} 
& 0 & 5 & 3 & 3 & 11 & 5.50 & 4 & 14 \\
& & 2 & 4 & 3 & 3 & 12 & 5.50 & 5 & 14 \\
& \multirow{2}{*}{Deduction} & 0 & 4 & 0 & 0 & 2 & 1.50 & 2 & 9 \\
& & 4 & 1 & 0 & 1 & 0 & 0.50 & 0 & 9 \\
\midrule
\multirow{6}{*}{\rotatebox{90}{Gemma3-4B}}
& \multirow{2}{*}{AR-LSAT} & 0 & 9 & 0 & 1 & 5 & 3.75 & 1 & 24 \\
& & 3 & 0 & 0 & 0 & 5 & 1.25 & 2 & 24 \\
& \multirow{2}{*}{FOLIO} & 0 & 11 & 11 & 10 & 14 & 11.50 & 11 & 14 \\
& & 2 & 10 & 5 & 10 & 14 & 9.75 & 9 & 14 \\
& \multirow{2}{*}{Deduction} & 0 & 3 & 1 & 1 & 6 & 2.75 & 3 & 9 \\
& & 4 & 6 & 0 & 1 & 8 & 3.75 & 6 & 9 \\
\midrule
\multirow{6}{*}{\rotatebox{90}{Gemma3-12B}}
& \multirow{2}{*}{AR-LSAT} & 0 & 16 & 2 & 6 & 16 & 10.00 & 9 & 24 \\
& & 3 & 7 & 0 & 0 & 4 & 2.75 & 4 & 24 \\
& \multirow{2}{*}{FOLIO} & 0 & 11 & 8 & 12 & 13 & 11.00 & 9 & 14 \\
& & 2 & 13 & 9 & 12 & 13 & 11.75 & 12 & 14 \\
& \multirow{2}{*}{Deduction} & 0 & 6 & 4 & 6 & 7 & 5.75 & 6 & 9 \\
& & 4 & 6 & 4 & 5 & 8 & 5.75 & 8 & 9 \\
\midrule
\multirow{6}{*}{\rotatebox{90}{Gemma3-27B}}
& \multirow{2}{*}{AR-LSAT} & 0 & 12 & 2 & 5 & 6 & 6.25 & 6 & 24 \\
& & 3 & 16 & 3 & 5 & 10 & 8.50 & 10 & 24 \\
& \multirow{2}{*}{FOLIO} & 0 & 13 & 11 & 13 & 13 & 12.50 & 12 & 14 \\
& & 2 & 12 & 10 & 10 & 14 & 11.50 & 12 & 14 \\
& \multirow{2}{*}{Deduction} & 0 & 6 & 5 & 6 & 9 & 6.50 & 8 & 9 \\
& & 4 & 7 & 5 & 9 & 9 & 7.50 & 9 & 9 \\
\midrule
\multirow{6}{*}{\rotatebox{90}{Llama3-8B}}
& \multirow{2}{*}{AR-LSAT} & 0 & 5 & 0 & 0 & 2 & 1.75 & 7 & 24 \\
& & 3 & 1 & 0 & 0 & 3 & 1.00 & 5 & 24 \\
& \multirow{2}{*}{FOLIO} & 0 & 8 & 2 & 4 & 10 & 6.00 & 7 & 14 \\
& & 2 & 6 & 3 & 5 & 8 & 5.50 & 7 & 14 \\
& \multirow{2}{*}{Deduction} & 0 & 5 & 3 & 5 & 6 & 4.75 & 6 & 9 \\
& & 4 & 0 & 0 & 0 & 3 & 0.75 & 3 & 9 \\
\midrule
\multirow{6}{*}{\rotatebox{90}{Qwen3-4B}}
& \multirow{2}{*}{AR-LSAT} & 0 & 20 & 18 & 22 & 21 & 20.25 & 21 & 24 \\
& & 3 & 18 & 14 & 17 & 15 & 16.00 & 18 & 24 \\
& \multirow{2}{*}{FOLIO} & 0 & 13 & 12 & 13 & 14 & 13.00 & 11 & 14 \\
& & 2 & 12 & 12 & 13 & 12 & 12.25 & 10 & 14 \\
& \multirow{2}{*}{Deduction} & 0 & 8 & 8 & 9 & 8 & 8.25 & 9 & 9 \\
& & 4 & 7 & 6 & 8 & 8 & 7.25 & 8 & 9 \\
\midrule
\multirow{6}{*}{\rotatebox{90}{Phi4mini-4B}}
& \multirow{2}{*}{AR-LSAT} & 0 & 6 & 0 & 2 & 6 & 3.50 & 2 & 24 \\
& & 3 & 6 & 0 & 0 & 9 & 3.75 & 4 & 24 \\
& \multirow{2}{*}{FOLIO} & 0 & 12 & 7 & 10 & 12 & 10.25 & 10 & 14 \\
& & 2 & 8 & 2 & 5 & 12 & 6.75 & 9 & 14 \\
& \multirow{2}{*}{Deduction} & 0 & 6 & 2 & 2 & 5 & 3.75 & 5 & 9 \\
& & 4 & 5 & 1 & 2 & 4 & 3.00 & 1 & 9 \\
\bottomrule
\end{tabular}
\caption{Full qualitative evaluation results across models, datasets, and label-preserving edit levels. Overall reasoning and Correct final outputs are reported and should be considered with respect to the total number of records used in the analysis, termed as Count. For presentation, edit levels are limited to level 0 and the maximum available edit level for each dataset.}
\vspace{-2mm}
\label{tab:appendix-full-qualitative}
\end{table}

%% file: norm_logic/singleedits_full.tex
\begin{table*}[ht]
\centering
\small
\resizebox{\textwidth}{!}{%
\begin{tabular}{llrrrrrrrr}
\toprule
\multirow{2}{*}{\textbf{}} 
& \multirow{2}{*}{\textbf{Model}} 
& \multicolumn{4}{c}{\textbf{Preserved Label}} 
& \multicolumn{4}{c}{\textbf{Altered Label}} \\
\cmidrule(lr){3-6} \cmidrule(lr){7-10}
& 
& \textbf{Base\#} & \textbf{Edit\#} & \textbf{Base Acc.} & \textbf{Edit Acc.}
& \textbf{Base\#} & \textbf{Edit\#} & \textbf{Base Acc.} & \textbf{Edit Acc.} \\
\midrule

\multirow{7}{*}{\rotatebox{90}{FOLIO}}
& \texttt{Gemma3-1B}      & 68 & 182 & 47.06\% & 41.47\% & 50 & 69 & 54.00\% & 26.67\% \\
& \texttt{Gemma3-4B}      & 68 & 182 & 72.06\% & 74.44\% & 50 & 69 & 72.00\% & 42.67\% \\
& \texttt{Gemma3-12B}     & 68 & 182 & 79.41\% & 85.34\% & 50 & 69 & 74.00\% & 58.00\% \\
& \texttt{Gemma3-27B}     & 68 & 182 & 76.47\% & 80.02\% & 50 & 69 & 80.00\% & 56.67\% \\
& \texttt{Llama3-8B}      & 68 & 182 & 52.94\% & 58.23\% & 50 & 69 & 54.00\% & 32.67\% \\
& \texttt{Qwen3-4B}       & 68 & 182 & 88.24\% & 88.11\% & 50 & 69 & 94.00\% & 71.33\% \\
& \texttt{Phi4mini-4B}   & 68 & 182 & 70.59\% & 70.12\% & 50 & 69 & 58.00\% & 53.33\% \\

\midrule

\multirow{7}{*}{\rotatebox{90}{Deduction}}
& \texttt{Gemma3-1B}      & 123 & 397 & 33.33\% & 29.34\% & 157 & 175 & 18.47\% & 21.34\% \\
& \texttt{Gemma3-4B}      & 123 & 397 & 46.34\% & 43.43\% & 157 & 175 & 31.21\% & 30.89\% \\
& \texttt{Gemma3-12B}     & 123 & 397 & 85.37\% & 88.77\% & 157 & 175 & 80.25\% & 72.56\% \\
& \texttt{Gemma3-27B}     & 123 & 397 & 86.99\% & 89.78\% & 157 & 175 & 87.26\% & 74.31\% \\
& \texttt{Llama3-8B}      & 123 & 397 & 69.11\% & 65.05\% & 157 & 175 & 49.68\% & 41.03\% \\
& \texttt{Qwen3-4B}       & 123 & 397 & 97.56\% & 96.88\% & 157 & 175 & 95.54\% & 95.22\% \\
& \texttt{Phi4mini-4B}   & 123 & 397 & 68.29\% & 71.42\% & 157 & 175 & 65.61\% & 57.38\% \\
\bottomrule
\end{tabular}
}
\caption{Base and edit accuracy across datasets and models under label-preserving and label-altering conditions. Accuracy values are reported as percentages with eval accuracy reported as a grouped accuracy.}
\label{tab:singleedits-all}
\end{table*}

%% file: norm_logic/backtranslation-prompts.tex
\begin{figure*}[ht]
    \centering
    \begin{tcolorbox}[
        width=\textwidth,
        colback=yellow!8,
        colframe=blue!45!black,
        title=AR-LSAT: Prompt and an In-Context Example for back-translation,
        fonttitle=\bfseries
    ]
    \underline{\textbf{System:}} \\
    You are given a symbolic logic expression and a natural-language statement with an incorrect logical operator. Correct the statement to match the symbolic logic. Always output in valid JSON only. Do not add any explanations or text outside the JSON. \\
    \noindent\rule{\textwidth}{0.4pt}\\
    \underline{\textbf{User:}} \\
    \textbf{Task Description:} \\
    You are given:
1) A symbolic logic expression describing a rule.
2) A natural-language statement that uses the correct entities and vocabulary but contains an incorrect logical operator.

The incorrect statement differs from the symbolic logic only in the logical operator used.
\noindent\rule{\textwidth}{0.4pt}\\
Important Logical Semantics:

- AND → both conditions must be true

- OR → at least one condition must be true

- XOR → exactly one condition is true (but not both)

- NOT → the condition is false \\
Output Requirements:...

Guidelines: ...\\
    \noindent\rule{\textwidth}{0.4pt}\\
    \underline{\textbf{Example:}} \\
\texttt{\{"symbolic\_logic":"Xor(position(L) == 8, position(H) == 8)",}\\
\texttt{"incorrect\_statement":"The eighth composition performed is either L or H.",}\\
\texttt{"correct\_statement":"The eighth composition performed is either L or H, but not both."\}}
    \end{tcolorbox}
    \caption{Instruction prompt (shortened) for translating first-order logic to natural-language statements for AR-LSAT}
    \label{fig:logic-correction-icl}
\end{figure*}

\begin{figure*}[ht]
    \centering
    \begin{tcolorbox}[
        width=\textwidth,
        colback=yellow!8,
        colframe=blue!45!black,
        title=Logical Deduction: Prompt and an In-Context Example for back-translation,
        fonttitle=\bfseries
    ]
    \underline{\textbf{System:}} \\
    You are a parser that converts logical expressions into natural-language statements... \\
    \noindent\rule{\textwidth}{0.4pt}\\
    \underline{\textbf{User:}} \\
    \textbf{Task Description:} \\
    You are given:
1) A symbolic logic expression describing a relationship (e.g., ordering, position, equality, or inequality) among objects in a domain.
2) A natural-language statement that uses the correct vocabulary but represents the logic incorrectly.
Your task is to interpret the symbolic logic expression and produce a correct natural-language statement that accurately reflects its logical meaning.\\
Output Requirements:...\\
Guidelines:...\\
\noindent\rule{\textwidth}{0.4pt}\\
    \underline{\textbf{Example:}} \\
\texttt{\{"symbolic\_logic":"tractor < bus",}\\
\texttt{"incorrect\_statement":"The tractor is older than the bus.",}\\
\texttt{"correct\_statement":"The tractor is younger than the bus."\}} \\
    \end{tcolorbox}
    \caption{Instruction prompt (shortened) for translating logical form to natural-language statements for Deduction}
    \label{fig:logic-correction-icl-deduction}
\end{figure*}

\begin{figure*}[ht]
    \centering
    \begin{tcolorbox}[
        width=\textwidth,
        colback=yellow!8,
        colframe=blue!45!black,
        title=FOLIO: Prompt and an In-Context Example for back-translation,
        fonttitle=\bfseries
    ]
    \underline{\textbf{System:}} \\
    You are a parser that converts First-Order Logic (FOL) expressions into natural-language statements... \\
    \noindent\rule{\textwidth}{0.4pt}\\
    \underline{\textbf{User:}} \\
    \textbf{Task Description:} \\
    You are given:\\
1) A first-order logic (FOL) expression.\\
2) A natural-language statement that uses the correct vocabulary but has an incorrect logical structure.\\
Your task is to interpret the FOL expression and produce the correct natural-language statement that accurately reflects its logical meaning.\\
Output Requirements:...\\
Guidelines:...\\
\noindent\rule{\textwidth}{0.4pt}\\
    \underline{\textbf{Example:}} \\
\texttt{\{"symbolic\_logic":"$\forall{x} (Fish(x) \rightarrow Plant(x))$",}\\
\texttt{"incorrect\_statement":"No fish are plants.",}\\
\texttt{"correct\_statement":"All fish are plants."\}} \\
    \end{tcolorbox}
    \caption{Instruction prompt (shortened) for translating logical form to natural-language statements for FOLIO}
    \label{fig:logic-correction-icl-folio}
\end{figure*}

%% file: norm_logic/qualitative-prompt.tex
\begin{figure*}[ht]
    \centering
    \begin{tcolorbox}[
        width=\textwidth,
        colback=yellow!8,
        colframe=blue!45!black,
        title=Reasoning Evaluation: Prompt for Qualitative Binary Coding,
        fonttitle=\bfseries
    ]
    \underline{\textbf{System:}} \\
    You are an expert qualitative-analysis system designed to classify patterns, strategies, and evaluation dimensions present in model-generated reasoning or text. You must always output a single valid JSON object...\\
    \noindent\rule{\textwidth}{0.4pt}\\
    \underline{\textbf{User:}} \\
    \textbf{Task Description:} \\
    Your task is to evaluate a reasoning path using the original task information and the model's reasoning text. The goal is to extract structured qualitative results from the reasoning path using four binary coding dimensions.\\

    You will be given:\\
    1) The original question, including any context, options, or constraints.\\
    2) The reasoning text produced by a model.\\

    Evaluate the reasoning using the following four codes. Assign each code a binary score, where \texttt{1} means the reasoning satisfies the code and \texttt{0} means it does not satisfy the code, has meaningful issues, ...\\

    \textbf{Evaluation Codes:}\\
    1) \textbf{Premise fidelity}: Assess whether the reasoning accurately uses only the information provided in the original question and avoids misreading, distorting, or inventing premises.\\
    2) \textbf{Inference validity}: Assess whether the reasoning steps logically follow from the given premises and from prior reasoning steps.\\
    3) \textbf{Completeness}: Assess whether the reasoning includes the necessary intermediate steps needed to support the conclusion.\\
    4) \textbf{Answer alignment}: Assess whether the final answer follows from and is supported by the reasoning.\\
    {Evaluation Rules:}...\\
    \noindent\rule{\textwidth}{0.4pt}\\
    \underline{\textbf{Assistant:}} \\
\begin{verbatim}
{
  premise_fidelity: {
    score: 0,
    justification: The model significantly reinterprets premise 5 ...
  },
  inference_validity: {
    score: 1,
    justification: 
  },
  ...
  }
}
\end{verbatim}

    \end{tcolorbox}
    \caption{Instruction prompt (shortened) for evaluating model-generated reasoning paths using four binary qualitative coding dimensions}
    \label{fig:qualitative-code-prompt}
\end{figure*}